\documentclass{esannV2}
\usepackage[dvips]{graphicx}
\usepackage[latin1]{inputenc}
\usepackage{amssymb,amsmath,array}
\usepackage{algorithm}
\usepackage{algorithmic}
\usepackage{subfigure}

\newcommand{\bfx}{{\textbf{x}}}

\newcommand{\bfu}{{\textbf{u}}}
\newcommand{\bfw}{{\textbf{w}}}
\newcommand{\bfy}{{\textbf{y}}}
\newcommand{\bfg}{{\textbf{g}}}

\newcommand{\bfdelta}{{\boldsymbol{\delta}}}
\newcommand{\bfepsilon}{{\boldsymbol{\epsilon}}}

\begin{document}
\title{Learning convolutional neural network to maximize Pos@Top performance measure}

\author{Yanyan Geng$^1$, Ru-Ze Liang$^2$
\footnote{Yanyan Geng and Ru-Ze Liang are the co-first authors of this paper.}
, Weizhi Li$^3$, Jingbin Wang$^4$,\\
Gaoyuan Liang$^5$, Chenhao Xu $^6$, Jing-Yan Wang$^{1}$,
\footnote{The study was supported by the open research program of the Provincial Key Laboratory for Computer Information Processing Technology, Soochow University, China (Grant No. KJS1324).}
\\
1 - Provincial Key Laboratory for Computer Information Processing Technology,\\
Soochow University, Suzhou 215006, China\\
Email: yanyangeng@outlook.com\\
2 - King Abdullah University of Science and Technology, Thuwal 23955, Saudi Arabia\\
3 - Suning Commerce R\&D Center USA, Inc., Palo Alto, CA 94304, United States\\
4 - Information Technology Service Center, \\
Intermediate People's Court of Linyi City, Linyi, China\\
Email: jingbinwang1@outlook.com\\
5 - Jiangsu University of Technology, Jiangsu 213001, China\\
6 - New York University Abu Dhabi, Abu Dhabi, United Arab Emirates\\
}

\maketitle

\begin{abstract}
In the machine learning problems, the performance measure is used to evaluate the machine learning models. Recently, the number positive data points ranked at the top positions (Pos@Top) has been a popular performance measure in the machine learning community. In this paper, we propose to learn a convolutional neural network (CNN) model to maximize the Pos@Top performance measure. The CNN model is used to represent the multi-instance data point, and a classifier function is used to predict the label from the its CNN representation. We propose to minimize the loss function of Pos@Top over a training set to learn the filters of CNN and the classifier parameter. The classifier parameter vector is solved by the Lagrange multiplier method, and the filters are updated by the gradient descent method alternately in an iterative algorithm. Experiments over benchmark data sets show that the proposed method outperforms the state-of-the-art Pos@Top maximization methods.
\end{abstract}

\section{Introduction}

In the machine learning and data mining applications, the performance measures are used to evaluate the performance of the predictive models \cite{liang2016novel,liang2016optimizing,fan2016stochastic,li2016nuclear}.
The popular performance measures include the area under the receiver operating characteristic curve (AUC), the recall-precision break-even point (RPB), the top $k$-rank precision (Top$k$ Pre), and the positives at top (Pos@Top) \cite{joachims2005support,li2016nuclear}. Recently, the performance of Pos@Top is being more and more popular in the machine learning community. This performance measures only counts the positive instances ranked before the first-ranked negative instance. The rate of these positive instances in all the positive instances is defined as the \textbf{Pos@Top}. In many machine learning applications, we observed that the top-ranked instances/classes plays critical roles in the performance evaluation, and the Pos@Top performance measure can give a good description about how the top-ranked instances/classes distribute \cite{Agarwal2011839,Boyd2012953,Li20141502}. Moreover, it is parameter free, and it use the top-ranked negative as the boundary of the recognized positive instance pool.
A few works were proposed to optimize the Pos@Top measure in the training process directly \cite{Li20141502,Boyd2012953}.
Among these three existing works of optimization of the Pos@Top, all the predictive models are linear models, they are designed to tickle the single-instance data, and the performance is still not satisfying.

To solve these problems, we develop a convolutional neural network (CNN) model to optimize the Pos@Top performance measure. This model is designed to tickle the multiple instance sequence as input.
We propose to learn the parameters of the CNN and classifier model, including the filters and the classifier parameter to maximize the Pos@Top.
We define a hinge loss function for this problem, and solve this problem by alternate optimization problem and develop an iterative algorithm.

\section{Proposed method}
\label{sec:method}

\subsection{Problem modeling}

The training set is composed of $n$ data points, and denoted as $\{(X_i,y_i)\}_{i=1}^n$, where $X_i = [\bfx_{i1},\cdots,\bfx_{im_{i}}]\in \mathbb{R}^{d\times m_i}$ is the input data of the $i$-th data point, $\bfx_{i\kappa}\in \mathbb{R}^d$ is the $d$-dimensional feature vector of the $\kappa$-th instance, and $y_i\in \{+1,-1\}$ is its binary class label.
To represent the $i$-th data point, we use a CNN model,
$
\bfg(X_i) = \max(\phi(W^\top X_i))\in \mathbb{R}^m.
$
In this model, $W = [\bfw_1,\cdots,\bfw_m]\in \mathbb{R}^{d\times m}$ is the matrix of $m$ filters, and $\bfw_k\in \mathbb{R}^d$ is the $k$-th filter vector. $\phi(\cdot)$ is a element-wise non-linear activation function.
$\max(\cdot)$ is the row-wise maximization operator.
To approximate the class label $y_i$ of a data point $X_i$ from its CNN representation $\bfg(X_i)$, we propose to use a linear classifier,
$
y_i \leftarrow f(X_i) = \bfu^\top \bfg(X_i)
$, where $\bfu\in \mathbb{R}^m$ is a parameter vector of the linear classifier.

The argued performance measure, Pos@Top, is defined as number of positive data points which are ranked before the top-ranked negative data point, $\max_{j:y_j=-1} f(X_j)$.
To maximize the Pos@Top, we argue that for any positive data point, its classification score should be larger than that of the top-ranked negative plus a margin amount,
\begin{equation}
\label{equ:condition}
\begin{aligned}
f(X_i) > \max_{j:y_j=-1} f(X_j) + 1, \forall~i:y_i=+1,
\end{aligned}
\end{equation}
we further propose a hinge loss function as follows to give a loss when this condition does not hold,
\begin{equation}
\label{equ:loss1}
\begin{aligned}
&\ell_{hinge}(X_i,y_i) = \max\left(0,\max_{j:y_j=-1} f(X_j) - f(X_i) + 1\right ),~\forall~i:y_i=+1.
\end{aligned}
\end{equation}
To learn the CNN classifier parameters $W$ and $\bfu$ to maximize the Pos@Top, we should minimize the loss function of (\ref{equ:loss1}) over all the positive data points. Meanwhile we also propose to regularize the filter parameters and the full connection weights to prevent the over-fitting problem, and the squared $\ell_2$ norms of $\bfu$ and $W$ are minimized.
The overall objective function of the learning problem is given as follows,
\begin{equation}
\label{equ:objective2}
\begin{aligned}
&\underset{W, \bfu, \xi_i|_{i,y_i=+1},\theta}{\min}~
\left \{\frac{1}{2}\|\bfu\|_2^2+ C_1 \sum_{i:y_i=+1} \xi_i
+C_2\|W\|_2^2
\right \},\\
&s.t.~
\forall~j,y_j=-1:~f(X_j)\leq \theta. \forall~i,y_i=+1:0\leq \xi_i,~and~\theta - f(X_i) + 1 \leq \xi_i,
\end{aligned}
\end{equation}
where $C_1$ and $C_2$ are the tradeoff parameters, $\theta = \max_{j:y_j=-1} f(X_j)$, and $\xi_i = \max \left (0, \max_{j:y_j=-1} f(X_j) - f(X_i) + 1 \right)
=\max \left (0, \theta - f(X_i) + 1 \right)$.

\subsection{Problem optimization}
\label{sec:opt}

The dual form of the optimization problem is as follows,
\begin{equation}
\begin{aligned}
\max_{\bfdelta}&
\min_{W}
\left \{ -\frac{1}{2} \sum_{i,i'=1}^n \delta_i \delta_{i'} y_i y_{i'} \bfg(X_i)^\top \bfg(X_{i'})
\right.
+C_2\|W\|_2^2 + \bfepsilon^\top\bfdelta\\
&=
-\frac{1}{2} \sum_{i,i'=1}^n \delta_i \delta_{i'} y_i y_{i'} \sum_{k=1}^m \phi(\bfw_k^\top \bfx_{i\psi_{ik}}) \phi(\bfw_k^\top \bfx_{i'\psi_{i'k}})
\left.
+C_2\|W\|_2^2 + \bfepsilon^\top\bfdelta
\vphantom{\sum_1^{1^1}}
\right\},\\
s.t.
&\bfdelta\geq 0, diag(\bfepsilon)\bfdelta  \leq  C_1 \bf1, \bfy^\top\bfdelta = 0,
\end{aligned}
\end{equation}
where $\bfepsilon = [\epsilon_1,\cdots,\epsilon_n]\in\{1,0\}^n$ and $\epsilon_i=1$ if $y_i=1$, and $0$ otherwise. $\delta_i$ is the Lagrange multiplier variable for the constraint $\theta - f(X_i) + 1 \leq \xi_i$ if $y_i = +1$, and that for the constraint $f(X_i)\leq \theta$ otherwise. $\bfdelta = [\delta_1,\cdots,\delta_n]^\top$, $\bfy = [y_1,\cdots,y_n]^\top$. $\psi_{ik}= {\arg\max}_{\kappa=1}^{m_i} \phi(\bfw_k^\top\bfx_{i\kappa})$,
and it indicates the instance in a bag which gives the maximum response with regard to a filter. We propose to use an alternate optimization strategy to optimize this problem. In an iterative algorithm, $\bfdelta$ and $W$ are updated alternately. The iterative algorithm is named as ConvMPT.

\subsubsection{Updating $\bfdelta$}

To optimize $\bfdelta$, we fix $W$ and remove the irrelevant term to obtain the following optimization problem,
\begin{equation}
\begin{aligned}
\max_{\bfdelta}&
\left \{
-\frac{1}{2} \sum_{i,i'=1}^n \delta_i \delta_{i'} y_i y_{i'} \sum_{k=1}^m \phi(\bfw_k \bfx_{i\psi_{ik}}) \phi(\bfw_k \bfx_{i'\psi_{i'k}})
\right.
\left.
+ \bfepsilon^\top\bfdelta
\vphantom{-\frac{1}{2} \sum_{i,i'=1}^n}
\right\},\\
s.t.
&\bfdelta\geq 0, diag(\bfepsilon)\bfdelta  \leq  C_1 \bf1, \bfy^\top\bfdelta = 0.
\end{aligned}
\end{equation}
This is a linear constrained quadratic programming problem, and we can use an active set algorithm to solve it.

\subsubsection{Updating $W$}

To optimize $W$, we fixe $\bfdelta$ and remove the irrelevant terms to obtain the following problem,
\begin{equation}
\begin{aligned}
\min_{W}
&\left \{
-\frac{1}{2} \sum_{i,i'=1}^n \delta_i \delta_{i'} y_i y_{i'} \sum_{k=1}^m \phi(\bfw_k^\top \bfx_{i\psi_{ik}}) \phi(\bfw_k^\top \bfx_{i'\psi_{i'k}})
\right.
\left. +C_2\|W\|_2^2 = \sum_{k=1}^m s(\bfw_k)
\vphantom{\frac{1}{2} \sum_{i,i'=1}^n}
\right\},
\end{aligned}
\end{equation}
where
$s(\bfw_k) =
-\frac{1}{2} \sum_{i,i'=1}^n \delta_i \delta_{i'} y_i y_{i'} \phi(\bfw_k^\top \bfx_{i\psi_{ik}}) \phi(\bfw_k^\top \bfx_{i'\psi_{i'k}})
+C_2\|\bfw_k\|_2^2$.
It is clear that $s(\bfw_k)$ is an independent function of $\bfw_k$, thus we can update the filters one by one. When one filter is being updated, other filters are fixed. The updating of $\bfw_k$ is conducted by gradient descent,
$
\bfw_k \leftarrow \bfw_k - \eta\nabla s_{\bfw_k}(\bfw_k),
$
where $\nabla s_{\bfw_k}(\bfw_k)$ is the gradient function of $s_{\bfw_k}(\bfw_k)$.

\section{Experiments}
\label{sec:exp}

In this section of experiments, we evaluate the proposed algorithm over several multiple instance data set for the problem of maximization of Pos@Top.

\subsection{Experimental data sets and setup}

In the experiment, we use tree types of data set --- image set, text set, and audio set.
The image set used by us is the Caltech-256 dataset. In this set, we have 30,607 images in total. These images belongs to 257 classes. Each image is presented as a bag of multiple instances. To this end, each image is split into a group of small image patches, and each patch is an instance.
The text data set used in this experiment is the Semeval-2010 task 8 data set. It contains 10,717 sentences, and these sentences belongs to 10 different classes of relations. Each sentence is composed of several words, and thus it is natural a multiple instance data set. Each word is represented as a feature vector of 100 dimensions using a word embedding algorithm.
The audio data set used in this experiment is the Spoken Arabic digits (SAD). In this data set, there are 8,800 sequences of voice signal. These voice signal sequences belongs to 10 classes of digits. To represent each sequence, we split it to a group of voice signal frames. Each frame is an instance, thus each sequence is a bag of multiple instances.
To perform the experiments over the data sets, we use the 10-fold cross-validation protocol to split training and test sets. The values of the tradeoff parameters of the ConvMPT are chosen by the cross-validation over the training set in the experiments, and the average Pos@Top values over the test sets are reported as the results. We use a one-vs-all strategy for the multi-class problem.

\subsection{Experimental results}

\begin{figure}[!htb]
\center
\subfigure[]{
\label{fig:fig1}
\includegraphics[width=.45\textwidth]{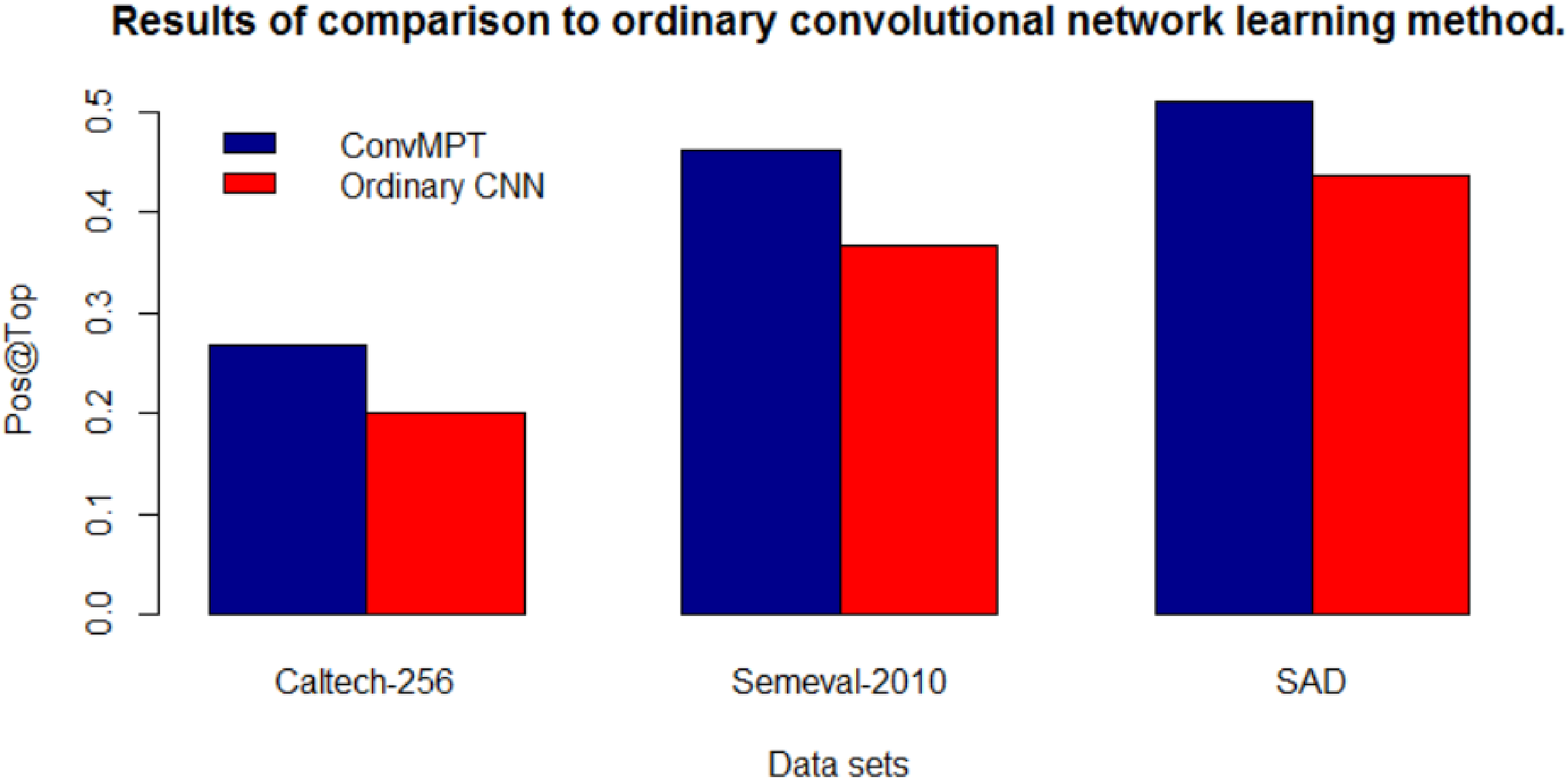}}
\subfigure[]{
\label{fig:fig2}
\includegraphics[width=.45\textwidth]{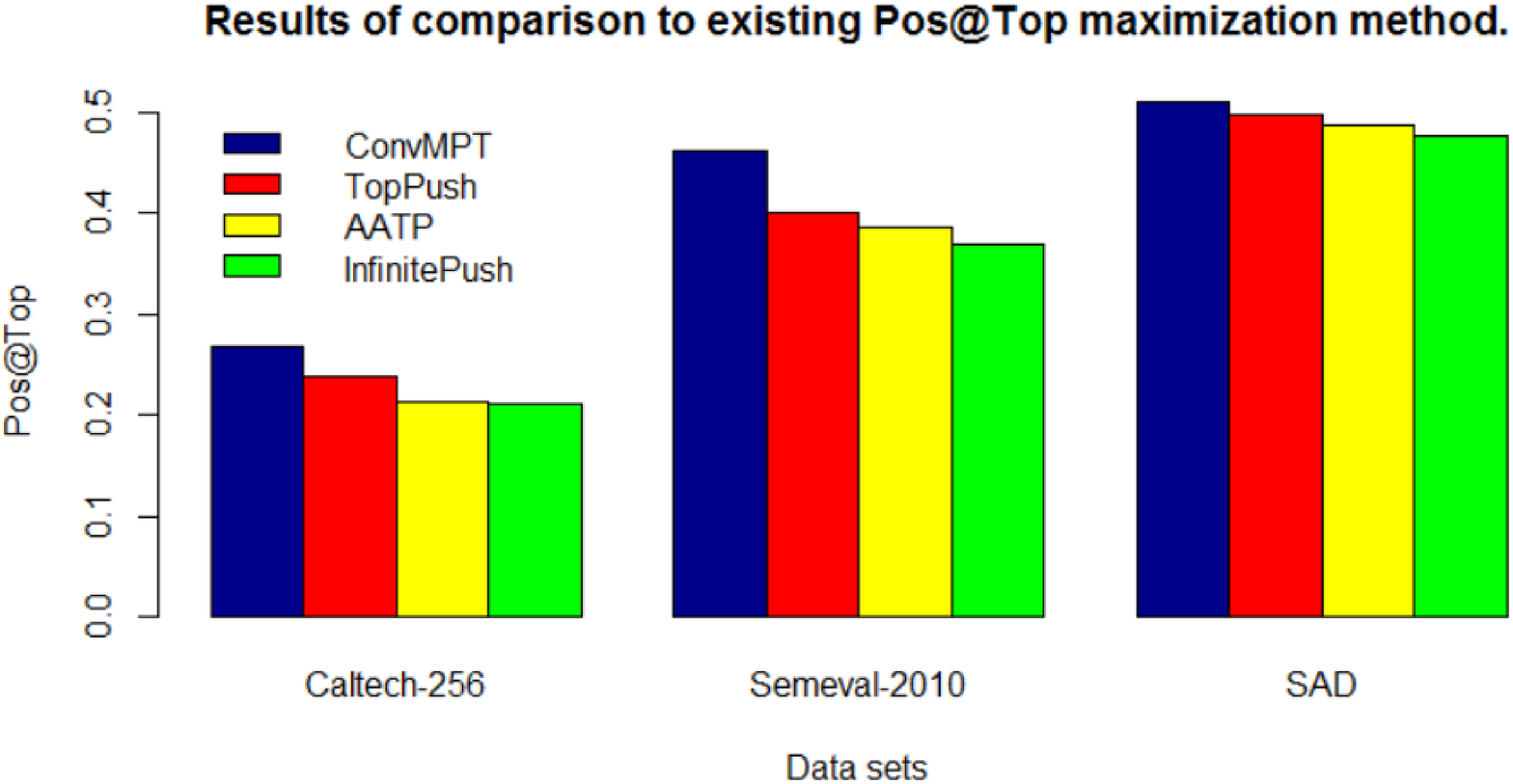}}
\caption{Results of comparison to ordinary convolutional network learning method and to existing Pos@Top maximization methods.}\label{fig:fig1}
\end{figure}

We firstly compare the proposed algorithm against ordinary convolutional network learning method. We use the TensorFlow as the implementation of the ordinary convolutional network model, and it logistic loss function as measure of performance measure. The results are reported in
Figure \ref{fig:fig1}. The results clearly show that the proposed method outperforms the ordinary CNN significantly over all the three data sets of different types.
We also compare the proposed method against some other algorithms for optimization of the Pos@Top performance. The compared methods are TopPush proposed by Li et al. \cite{Li20141502}, AATP proposed by Boyd et al. \cite{Boyd2012953}, and InfinitePush proposed by Agarwal \cite{Agarwal2011839}. The comparison results are shown in
Figure \ref{fig:fig2}. According to the results reported in the figure, the proposed convolutional Pos@Top maximizer outperforms the other methods. The compared methods, although used different optimization methods, but all aims to optimize a linear classifier, but our model has a convolutional structure. The results also show that convolutional network is a good choice for the problem of optimization of Pos@Top.

\section{Conclusion}
\label{sec:conclusion}

In this paper, we propose a novel model to maximize the performance of Pos@Top. The proposed model has a structure of CNN. The parameter learning of CNN is to optimize the loss function of Pos@Top. We propose a novel iterative learning algorithm to solve this problem. Meanwhile we also propose to minimize the squared $\ell_2$ norm of the filter matrix of the convolutional layer. The proposed algorithm is compared to the ordinary CNN and the existing Pos@Top minimization method, and the results show its advantage. In the future, we will apply the proposed method to application of computer vision \cite{shao2015face,tan2016robust}, sensor network \cite{cai20162,cai2016low}, medical imaging \cite{squiers2016multispectral,li2015burn,thatcher2016multispectral}, and material engineering \cite{wang2016donor,wolf2016benzo}.

\begin{footnotesize}

\end{footnotesize}

\end{document}